\documentclass{article}

\usepackage{arxiv}

\usepackage[utf8]{inputenc} 
\usepackage[T1]{fontenc}    
\usepackage{hyperref}       
\usepackage{url}            
\usepackage{booktabs}       
\usepackage{amsfonts}       
\usepackage{nicefrac}       
\usepackage{microtype}      
\usepackage{lipsum}		
\usepackage{graphicx}
\usepackage{natbib}
\usepackage{doi}
\usepackage{tcolorbox} 

\title{DataGpt-SQL-7B: An Open-Source Language Model for Text2SQL}

\author{ \hspace{1mm}Lixia Wu {$\ast$}\\
	Cainiao Network\\
	\thanks{\texttt{wallace.wulx@cainiao.com}} \\
        \And
        \hspace{1mm}Peng Li {$\ast$}\\
	Cainiao Network\\
	\texttt{lipeng.lipeng@cainiao.com} \\
	\And
        \hspace{1mm}Junhong Lou {$\ast$}\\
	Cainiao Network\\
	\texttt{loujunhong.loujunh@cainiao.com}\\
        \And
        \hspace{1mm}Lei Fu\\
	Cainiao Network\\
	\texttt{leo.ful@cainiao.com} 
}

\renewcommand{\shorttitle}{\textit{arXiv} Template}

\hypersetup{
pdftitle={A template for the arxiv style},
pdfsubject={q-bio.NC, q-bio.QM},
pdfauthor={David S.~Hippocampus, Elias D.~Striatum},
pdfkeywords={First keyword, Second keyword, More},
}

\begin{document}
\maketitle
\footnotetext[3]{\textsuperscript{$\dagger$}Corresponding authors.}
\begin{abstract}
In addressing the pivotal role of translating natural language queries into SQL commands, we propose a suite of compact, fine-tuned models and  self-refine mechanisms to democratize data access and analysis for non-expert users, mitigating risks associated with closed-source Large Language Models. Specifically, we constructed a dataset of over 20K sample for Text-to-SQL as well as the preference dateset, to improve the efficiency in the domain of SQL generation. To further ensure code validity, a code corrector was integrated into the model. Our system, DataGpt-sql, achieved 87.2\% accuracy on the spider-dev, respectively, showcasing the effectiveness of our solution in text-to-SQL conversion tasks. Our code, data, and models are available at \url{https://github.com/CainiaoTechAi/datagpt-sql-7b}


\end{abstract}


\section{Introduction}

The problem which translates natural language queries into Structured Query Language (SQL) pervades a multitude of real-world scenarios. An effective text-to-SQL methodology significantly enhances the efficiency of data analysis and retrieval, empowering non-expert users to interact with databases through their native language, democratizing access to and analysis of data. However, the reliance on powerful, close-sourced Large Language Models (LLM
) and sophisticated prompting techniques to address these tasks may introduce risks associated with openness, privacy, and substantial costs.

In light of these considerations, we propose a series of compact, fine-tuned models to tackle text-to-SQL problems. Schema linking, particularly the alignment of table and column information, is a recurrent issue in text-to-SQL tasks. We proposed the cross-DataBase and Inner-DataBase methods to construct samples, improving the performance of model to identify the correct schema and columns from context. To enhance the model's ability to generate  valid SQL and conform to specific code syntax, we create a preference dataset for text-to-SQL and utilize Direct Preference Optimization (DPO) for additional fine-tuning. Meanwhile, we leverage the reasoning capabilities of large language model to self-refine the generated SQL according to the result of executor. Our model and system have achieved competitive results on the spider-dev benchmarks, attaining 87.2\%(EX) and 83.5\% (TS) accuracy, respectively, while the pure model alone achieved 84.8\%(EX) and 81.5\%(TS) accuracy.




The remainder of this paper is organized as follows: Section 2 delves into related work on text-to-SQL. Section 3 provides a problem statement and details the implementation of our proposed model, DataGpt-sql. Section 4 presents the experimental outcomes, showcasing the efficacy of our approach. Finally, Section 5 gives the conclusion of this paper.
\section{Related Work}

\subsection{Text-to-SQL}
The task of generating accurate SQL queries from natural language questions, known as text-to-SQL, is a thriving research area that intersects the fields of natural language processing (NLP) and database management. Early contributions from the database community initiated significant advancements, such as the use of custom templates \cite{Zelle1996LearningTP}, but these approaches often required extensive manual effort.

Recent methodologies have increasingly integrated transformer-based models, notably sequence-to-sequence architectures \cite{Sutskever2014SequenceTS,Vaswani2017AttentionIA}. These models have demonstrated effectiveness in generating sequences from given inputs, making them well-suited for tasks like text-to-SQL \cite{qin2022surveytexttosqlparsingconcepts}. Early implementations, such as IRNet \cite{guo-etal-2019-towards}, employed a bidirectional Long Short-Term Memory (LSTM) architecture to encode queries while using self-attention mechanisms to comprehend database schema representations. Subsequent models, like RAT-SQL \cite{wang-etal-2020-rat} and \cite{qi-etal-2022-rasat}, enhanced this integration using relation-aware self-attention, while SADGA \cite{Cai_nips_21} and LGESQL \cite{cao2021lgesql} employed graph neural networks for relational structure modeling. Despite these advances, existing methods have yet to achieve execution accuracies above 80\% on challenging benchmarks like the Spider dataset \cite{yu-etal-2018-spider}.

Recent works leveraging Large Language Models (LLMs) have provided new momentum to the text-to-SQL domain. Initial strategies \cite{rajkumar2022evaluating} utilized zero-shot in-context learning for SQL query generation, which several subsequent models took further, including DIN-SQL \cite{pourreza2024din}, DAIL-SQL \cite{gao2023text}, and others, effectively applying methodologies like Chains of Thought (CoT) \cite{wei2022chain} and self-consistency \cite{wang2022self}. These innovative approaches reveal that task decomposition and example-based prompt engineering can significantly boost LLM performance, while efforts such as those in DAIL-SQL and other models propose fine-tuning strategies aimed at matching or surpassing the effectiveness of larger proprietary models. This category emphasizes the use of powerful closed-source base models, such as GPT-4 and its variations.

In contrast, this approach focuses on developing specialized models that engage in fine-tuning on specific text-to-SQL tasks, often using relatively smaller datasets. An example of this is the CodeS model series \cite{CodeS_2024}, which is an open-source language model explicitly designed for text-to-SQL tasks. It achieves significant gains in SQL generation and natural language understanding through incremental pre-training. It also incorporates a comprehensive database prompt construction strategy and bidirectional data augmentation methods, allowing the model to adapt flexibly to various database domains.

Using closed-source models raises concerns about privacy, openness, and cost. An alternative is to build domain-specific models tailored to specific tasks, which involves fine-tuning on a smaller scale within the domain. Another innovative approach treated the text-to-SQL as  as a sub-task of code generation, and train various domain-specific models. \cite{CodeS_2024} proposed an open-source language model series specifically designed for text-to-SQL tasks, which has achieved significant improvements in SQL generation and natural language understanding through incremental pre-training. It also adopts a comprehensive database prompt construction strategy and a bidirectional data augmentation method, allowing the model to adapt more flexibly to databases in different domains. SENSE \cite{yang2024synthesizing} introduced a novel data generation technique that combines the strengths of robust models like GPT-4 with weaker open-source models to create a hybrid training regimen. This method employs Direct Preference Optimization (DPO) to align model preferences effectively, enhancing performance while increasing data diversity.

\section{Methodology}
\subsection{Question Representation}
The objective of text-to-SQL is to generate a SQL query based on a natural language question $Q$ and a database $D$, such that the SQL query can be executed to answer the question:

\begin{equation}
    S = Parser(q,D)
\end{equation}
where the  $Parser()$ is designed to interpret the provided $Q$ using $D$  and produce $S$. contains database schema (i.e., tables and
columns) and database metadata, which contains column types,
comments, column values, primary keys, and foreign key relations.

\subsection{ DataGpt-SQL: domain-specific language model for Text-to-SQL}

\subsubsection{Data Augmentation for Supervised Fine-tuning}
To enhance the capabilities of existing language models in SQL generation and natural language understanding, we constructed a dataset of over 20,000 pairs for Text-to-SQL samples, and fine-tuned the CodeQwen1.5-7B-Chat \cite{bai2023qwen}. Specific prompt examples can be referenced in the following colorBox.
 Additionally, to further strengthen the model's ability to select tables and column names, we proposed the cross-DataBase and Inner-DataBase methods for sample construction.

The correct schema linking column selection has been shown to improve the performance of SQL generation. However, historically, most tasks have been accomplished through elaborate prompts or external retrieval systems that achieve the aforementioned functions. We often utilize constructed samples, allowing the model to automatically learn the ability to identify the correct schema and columns from environmental information.

Specifically, data augmentation consists of two approaches: Cross-DB augmentation and Inner-DB augmentation:

\textbf{Cross-DB Augmentation:}
\begin{enumerate}
    \item For a given sample, identify its Primary Key (Foreign Key) and scan other databases to obtain a set $Q$ that contains schema including that field.
    \item Randomly select $N$ schema from the set $Q$ (where $ 1 \leq N \leq 3 $) and insert them into the List of sample's Schema.
\end{enumerate}

\textbf{Inner-DB Augmentation:}
\begin{enumerate}
    \item Referencing the SFT sample construction method from CodeS\cite{CodeS_2024}, for a given sample, extract all table schema from the corresponding database. Within this schema list, randomly add or remove unused tables and columns based on the query. A sample retains a maximum of 6 tables, with each table holding a maximum of 10 columns.
\end{enumerate}

The goal of Cross-DB augmentation is to enhance the ability to select schema, whereas Inner-DB augmentation aims to improve the ability to choose the correct columns.

\begin{tcolorbox}
[colback=blue!5!white, colframe=blue!75!black, title=INPUT FOR Supervised Fine-tuning]
Database prompt:
CREATE TABLE stadium(Stadium\_ID, Location, Name,Capacity, Highest, Lowest, Average);\\ 
CREATE TABLE singer(Singer\_ID, Name, Country, Song\_Name, Song\_release\_year, Age, Is\_male);\\ 
CREATE TABLE concert(concert\_ID, concert\_Name, Theme, Stadium\_ID, Year); \\ 
CREATE TABLE singer\_in\_concert(concert\_ID, Singer\_ID). \\ 
-- Using valid SQLite, answer the following questions for the tables provided above. \\ 
-- How many singers do we have?
\end{tcolorbox}

\subsubsection{alignment with self-consistency executor result}
To further increase the probability of generating executable SQL and the ability to adhere to specific code syntax, we attempt to construct preference dataset for text-to-SQL and use DPO to further fine-tune the model. Specifically, common errors in generated SQL may include the following issues: Wrong Table Name, Wrong Column Name, and Missing quotation when encountering fields with special characters (such as space or slash). To minimize these occurrences, for the same Text2SQL problem, under a Temperature of 0.5, we generate N responses and execute all of them. Any responses that produce execution results inconsistent with the correct SQL execution results for that sample are considered rejected samples, while the correct SQL is selected as the chosen sample, where the following  text box exhibit one example of dateset in DPO training.

\begin{tcolorbox}
[colback=blue!5!white, colframe=blue!75!black, title=one example of dateset in DPO training]
\# prompt\\ 
CREATE TABLE tryout(pid, cname, decision, ppos);
CREATE TABLE player(hs, pname, ycard, pid);
CREATE TABLE college(cname, enr, state).
-- Using valid SQLite, answer the following questions for the tables provided above.

-- For each position, what is the minimum time students spent practicing?

\# chosen\\ 
SELECT min(player.hs) ,   tryout.ppos FROM tryout JOIN player ON tryout.pid  =  player.pid GROUP BY tryout.ppos

\# rejected\\ 
SELECT min(HS) ,  ppos FROM player GROUP BY ppos
\end{tcolorbox}

\subsubsection{One Efficient Reflection Agent based on Result of Executor}
\begin{figure}
    \centering
    \includegraphics[width=1.0\linewidth]{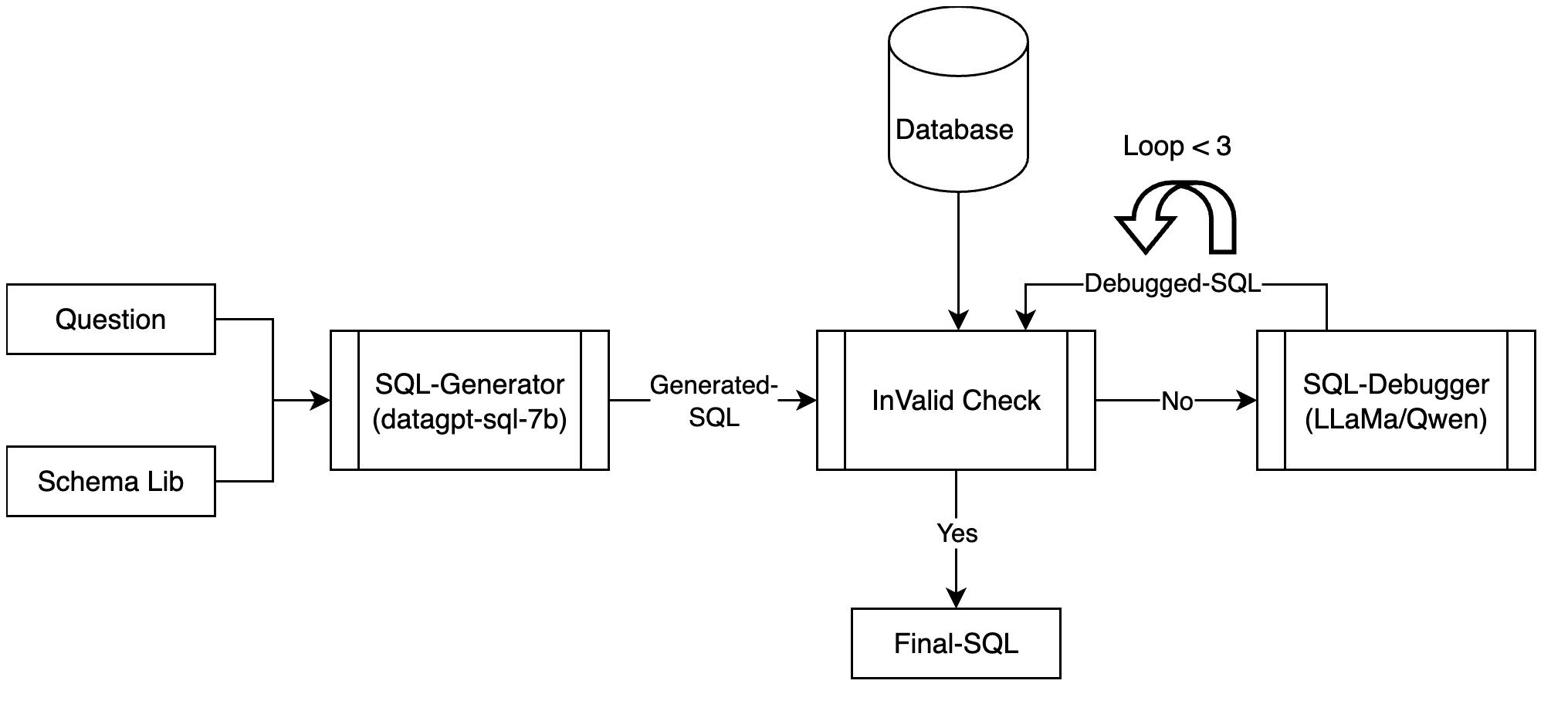}
    \caption{The pipeline for datatgpt-sql-7B + Invalid-SQL Feedback}
    \label{fig:enter-label}
\end{figure}

The flow diagram in \ref{fig:enter-label} illustrates the pipeline of integration DataGpt-sql with the mechanism of refine invalid-SQL, which named as "datatgpt-sql-7B+Invalid-SQL Feedback". 

The pipeline begins with a "Question" and a "Schema Lib" (which stands for a library containing the database schema). These inputs are essential for generating the SQL query.  The "SQL-Generator (DataGpt-sql-7B)" component takes the question and schema library as inputs and generates an SQL query. This SQL query is designed to retrieve the data specified by the question from the database. The generated SQL query is then passed through an "Invalid Check" process. This step checks the SQL query for errors or if it adheres to the specific code syntax. If the "Invalid Check" returns a "No," the process moves on to the "SQL-Debugger (LLaMa/Qwen)" component. This debugger is responsible for converting the SQL query into a "Debugged-SQL" format, which may include additional information or transformations to aid in debugging or understanding the query.  If the debugging process returns a "Yes," the flow moves to the "Final-SQL" box, indicating that the SQL query has been successfully debugged and is ready to be executed on the database.  Finally, the "Debugged-SQL" is sent to the "Database" for execution. Depending on the result of the SQL query, the process loop back to the "Invalid Check" components if there are issues or if additional debugging is required.

\section{Experiments}

\label{sec:others}

\subsection{Evaluation Metric}
For Spider-family benchmarks (including Spider, Spider-DK, Spider-Syn, Spider-Realistic, and Dr-Spider), we consider two prevalent evaluation metrics: execution
accuracy (EX) \cite{yu-etal-2018-spider} and test-suite accuracy (TS) \cite{zhong2020semantic}. The EX metric
evaluates whether the predicted and ground-truth SQL queries yield
the same execution results on the database. However, there can
be instances where EX gives false positives — situations where an
incorrect SQL prediction coincidentally produces the same output
as the correct query \cite{zhong2020semantic}. To counteract this, the test-suite accuracy
was developed. It assesses if the generated SQL query consistently
passes the EX evaluations across multiple database instances, which
are derived from automated database augmentations. Due to its
proficiency in reducing false positives, TS stands out as a more
trustworthy metric than EX. It’s worth noting that for Spider-DK
and Dr.Spider, the TS script lacks test suites for their respective
databases. As a result, we exclusively adopt the EX metric for them.
 The BIRD benchmark primarily relies on execution accuracy
(EX) as its evaluation metric, because the databases in BIRD typically contain a large number of values, minimizing the chances of
false positives. 




\subsection{Experimental results on Spider Benchmark}


In the experimental section of our research paper, we present a comparative analysis of various models designed for SQL generation tasks. The table below summarizes the performance of these models on two specific metrics: EX and TS. The proposed system, which incorporates feedback from invalid SQL queries during the training phase, outperforms all other models in terms of Execution Accuracy (EX) with a score of 87.2, and also achieves a high TS score of 83.8. Our proposed DataGpt-sql-7B model, without the additional feedback mechanism, still performs impressively with EX and TS scores of 84.8 and 81.5, respectively. Other models, such as DAIL-SQL + GPT-4 + Self-Consistency, SENCE-13B, and REDSQL-3B + NatSQL, also show strong performance, but they do not surpass our proposed  system. Notably, GPT-4, a general-purpose language model, lags significantly behind in both EX and TS metrics, highlighting the importance of specialized training for SQL generation tasks. Overall, these results underscore the effectiveness of our proposed "DataGpt-sql-7B+Invalid-SQL Feedback" in generating accurate SQL queries, outperforming a range of other models designed for similar tasks.

\begin{table}
	\caption{The experimental result on the benchmark of spider}
	\centering
 \begin{tabular}{lcc}
\toprule
Model & EX & TS \tabularnewline
\midrule
DataGpt-sql-7B + Invalid-SQL Feedback & 87.2 & 83.8 \tabularnewline
DataGpt-sql-7B & 84.8 & 81.5 \tabularnewline
DAIL-SQL + GPT-4 + Self-Consistency & 84.4 & - \tabularnewline
SENCE-13B & 84.1 & 83.5 \tabularnewline
REDSQL-3B + NatSQL & 84.1 & 73.5 \tabularnewline
DAIL-SQL + GPT-4 & 83.5 & 76.2 \tabularnewline
SENCE-7B & 83.2 & 81.7 \tabularnewline
Fine-tuned SQL-PaLM & 82.8 & 78.2 \tabularnewline
DIN-SQL + GPT-4 & 82.8 & 74.2 \tabularnewline
GPT-4 & 72.9 & 64.9 \tabularnewline
CodeQwen1.5-7B-Chat & 65.0 & 41.9 \tabularnewline
CodeLLaMa-7B-Instruct & 63.4 & 54.2 \tabularnewline
CodeLLaMa-13B-Instruct & 62.3 & 52.5 \tabularnewline
\bottomrule
\end{tabular}
	\label{tab:table}
\end{table}


\section{Conclusion}
This paper has introduced a novel approach to address the text-to-SQL challenge by proposing a series of compact, fine-tuned models that significantly enhance the efficiency of data analysis and retrieval for non-expert users. By leveraging high-quality datasets, schema linking techniques, and long-chain reasoning through stepwise Direct Preference Optimization, our model, Data-GPT-sql, demonstrates competitive performance on the spider-dev benchmarks, achieving 87.2\% (EX) and 83.5\% (TS) accuracy. The incorporation of a code corrector further minimizes the generation of invalid code, ensuring the reliability of the SQL commands produced. This work not only advances the field of text-to-SQL translation but also paves the way for more accessible and user-friendly database interaction, while mitigating the risks associated with proprietary models and prompting techniques. Future research will focus on expanding the model's capabilities in handling more complex database structures and improving its performance in multi-step reasoning tasks.

\bibliographystyle{unsrtnat}
\bibliography{references}  






\end{document}